\documentclass[a4paper, 10pt, conference]{ieeeconf}      

\usepackage{marvosym}
\usepackage{color}
\let\labelindent\relax
\usepackage[inline]{enumitem}
\usepackage{graphicx}
\usepackage{cite}
\usepackage{amsmath,amssymb,amsfonts}
\usepackage{algorithmic}
\usepackage{graphicx}
\usepackage{textcomp}
\usepackage{xcolor}
\usepackage{graphics} 
\usepackage{epsfig} 
\usepackage{mathptmx} 
\usepackage{times} 
\usepackage{soul}   

\usepackage{multirow}
\usepackage{multicol}
\usepackage{mathtools}
\usepackage{array}
\usepackage{xurl}

\usepackage{bbding}
\makeatletter
\let\NAT@parse\undefined
\makeatother
\usepackage[pagebackref,breaklinks,colorlinks,allcolors=blue]{hyperref}
\IEEEoverridecommandlockouts                              

\title{\bf Geometric Shortcuts for Complex Trunk Postures: Dual-Helicity Coupling Enables Low-Dimensional Control}

\author{Huishi Huang$^{1,2}$, Danlu Chen$^{1}$, Matteo Lo Preti$^{1}$,\\ Jun Liu$^{2}$\textsuperscript{\Letter}, Marcelo H. Ang Jr.$^{1}$$\textsuperscript{\Letter}$, and Cecilia Laschi$^{1}\textsuperscript{\Letter}$
\thanks{\textsuperscript{\Letter}Corresponding author: liuj@a-star.edu.sg, mpeangh@nus.edu.sg, mpeclc@nus.edu.sg}%
\thanks{$^{1}$Huishi Huang, Danlu Chen, Marcelo H. Ang Jr., and Cecilia Laschi are with the Advanced Robotics Centre (ARC) and with the Department of Mechanical Engineering, College of Design and Engineering, National
University of Singapore (NUS), 117608, Singapore.}%
\thanks{$^{2}$Huishi Huang and Jun Liu are with the Institute of High Performance Computing (IHPC), Agency for Science, Technology and Research (A*STAR), 138632, Singapore.}%
}
\begin{document}

\maketitle
\thispagestyle{empty}
\pagestyle{empty}

\begin{abstract}
How do elephant trunks generate complex postures without relying solely on fine segmental activation? We propose that part of this complexity arises from a low-dimensional geometric shortcut: dual-helicity coupling between opposite-handed oblique muscles. In a simplified soft-robotic prototype, varying only two geometric parameters generates a broad library of elephant-like postures, suggesting a dual-layer control architecture with implications for continuum robot design and biological hypotheses.

\end{abstract}

\section{Computational dimensionality in trunk control}
Among biological muscular hydrostats, the elephant trunk occupies a unique position: it combines the highest known muscle count with behavioral versatility spanning power grasping and precision manipulation. Elephant trunks display a striking repertoire of postures: bending, twisting, elongation and their combinations. Yet coordinating these movements presents a formidable computational challenge. Roughly 40,000 motor neurons in the facial nucleus innervate the entire elephant trunk \cite{kaufmann2022elephant} with 90,000 muscle fascicles. While motor units account for such anatomical ratios, coordinating a skeleton-free appendage composed of these fascicles still poses a high-dimensional control problem and substantial demands on neural bandwidth. Here we propose that part of this posture complexity may be reduced through a low-dimensional geometric shortcut, rather than relying entirely on high-dimensional local control to generate diverse, task-specific postures. 



It is thought that the physical ability of fine-grained movements is only achieved with sufficient muscular subdivision. Anatomical studies and muscle reconstruction have revealed the muscle composition of the elephant trunk with thousands of muscle fibers organized into many functional zones of fascicles \cite{longren2023dense, dagenais2021elephants}. Although the trunk musculature is anatomically segmented, it remains unclear whether the neural control is equivalently local. The classic segmental activation hypothesis requires the local muscle fibers be differentially activated in segments to form different curvature. However, it requires the individual muscle to be activated by independent neurons groups, which poses heavy burden to the neural load. 

In robotics, this mirrors challenges in continuum arms. The most common approach to generating complex shapes, including curvature reversals, is multi-segment actuation: each sign-change in curvature requires at least one independently controlled segment, so that manipulators capable of complex shapes demand a proportional number of actuators and corresponding control channels \cite{hannan2003kinematics, peng2025dexterous}. Beyond segmented actuation, reconfigurable designs offer an alternative: some achieve different curvatures by physically replacing or rearranging structural modules between tasks \cite{guan2025lattice, zhang2023preprogrammable}while others embed pre-programmable geometric parameters (e.g., tensegrity pretension) that can be adjusted in situ \cite{zhang2023situ, chen2025multimodal}.However, these approaches typically sacrifice generality: the designs are customized and difficult to reproduce, while each configuration is task-specific and hard to extend.



Is there a geometric shortcut? Classic understanding of elephants and common soft robot design rely too heavily on the segmented or modular principles for shape formation. In nature, the elephant trunk likely preserves its anatomical layout rather than reconfiguring discrete structural modules. Our observations hint at distributed low-dimensional coordination. We propose a dual-layer control strategy: layer 1 is the global coordination of the helical muscle system acting as an “integrator”, generating the gross posture; layer 2 utilizes the local segmental activation as a “differentiator” for fine manipulation such as local stiffening or grasping force.

In this view, elephants neither change the “hardware” nor rely solely on sophisticated individual muscle activation. To change the shape, the elephant trunk may modulate the effective anisotropy field generated by a fixed but deformable anatomical layout through selective muscle recruitment, thereby producing an effective helical transformation. A simplified soft-robotic prototype offers proof-of-concept of the hypothesis and maps from two geometric parameters to a library of elephant-like postures, while suggesting testable predictions for biology.

\section{Complex Shapes for Grasping}

Elephant trunks can in principle realize infinite DoFs, yet during grasping they appear to rely on a smaller repertoire of recurring posture families that reduce biomechanical complexity \cite{dagenais2021elephants}. These include distal bends, hooks, twist-and-bend combinations, S-bends, and more complex multi-node-bend configurations, selected according to object size, height, pose, and contact strategy.

Among these posture families, S-bends are particularly informative.  An S-bend, or reverse-bent posture, contains one or more \textbf{sign-changing curvatures} along the trunk. Such configurations can serve as a pre-shaping strategy during reaching, allowing the distal portion of the trunk to align with the target while limiting proximal bending and body displacement \cite{preti2023sensorized}. In this sense, S-bends expose a key control challenge: how can a highly compliant, continuously deformable appendage place and vary curvature reversals without relying entirely on concatenated segment-activated control?

\begin{figure*}[!t]
\centering
\includegraphics[width=\textwidth]{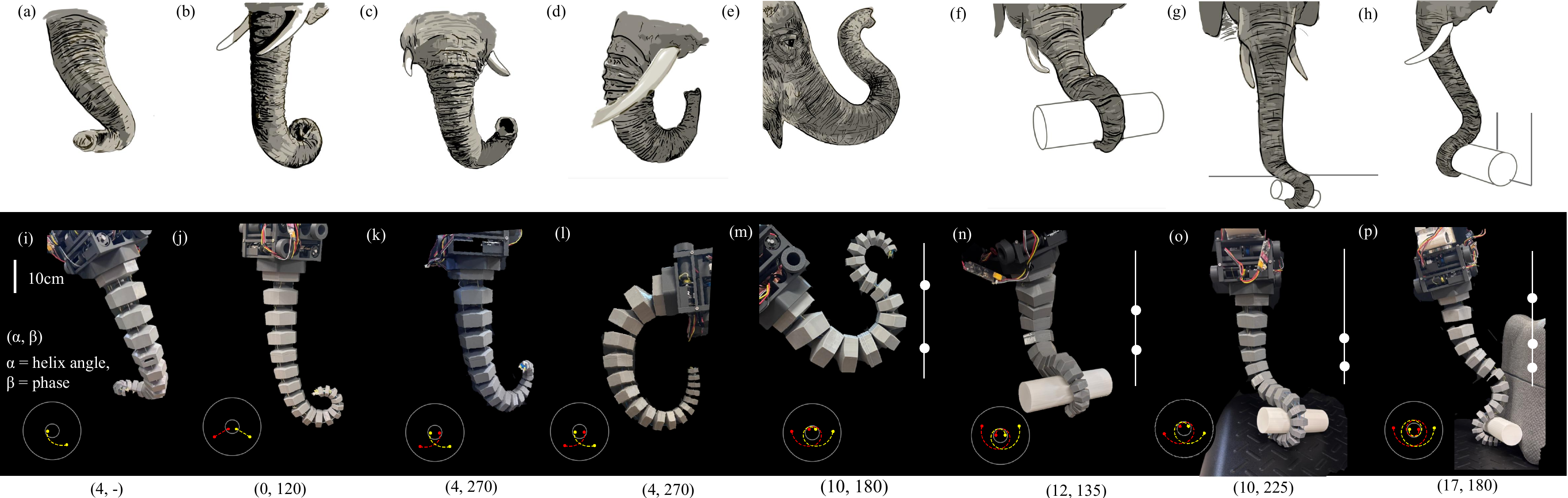} 
\caption{Dual-helicity geometry generates a library of elephant-like postures.
Top ($a-h$): representative trunk configurations observed in the field/literature, including twisting $(a)$, tight distal curls $(b)$, twist and bend $(c)$, single-curvature hooks $(d)$, S-shaped postures ($e-g$) and 3C-shaped $(h)$ with sign-changing curvature. Illustrations are manually drawn based on photo and videos: $(a)-(d)$ are  from \cite{leanza2024elephant}, $(e)$ from \cite{photosthai_elephant}, and $(f)-(h)$ from \cite{dagenais2021elephants}.  Bottom $(i-p)$: Our prototype reproduces these shapes by tuning the helix angle $\alpha$ and the relative phase $\beta$ between opposite-handed cable pairs. Insets show $\beta$ on a polar dial (color encodes handedness); numbers underneath indicate ($\alpha$, $\beta$) in degrees. White dots mark curvature sign-change nodes. A $10cm$ scale bar is shown in $(i)$. Noted that the robot base orientation sets the robots’ pose relative to gravity to match the photographed head pose. It does not change the intrinsic curvature profile.}
\label{method1}
\end{figure*}

Across tasks, the same underlying posture family can be expressed with different curvature magnitudes, distal engagement lengths, and node locations, from elevated reaches to large-object grasps and more complex multi-contact configurations (Fig.~\ref{method1} $A1-A8$). Taken together, these examples suggest that elephant grasping depends on a structured family of related shapes rather than a single canonical posture, with S-bends offering the clearest window into sign-changing curvature control.



\section{The dual-layer control Hypothesis}


\subsection{Dual-helicity as a geometric control principle}


Anatomy reveals longitudinal, oblique (helical), radial and transverse muscle systems arranged in discrete fascicular bands, wherein the helix muscles contribute predominantly to bending and torsion \cite{kier1985tongues, kier2012diversity}. We propose that a significant part of elephant trunk shape complexity can arise from the coupling of opposite-handed helical muscle groups. In our formulation, two geometric parameters govern the resulting posture family. The first is the helical obliquity $\alpha$, defined as the angle between the helical path and the trunk axis. The second is the phase span $\beta$, defined as the relative angular offset between opposite-handed helical muscle groups.

These two parameters play distinct roles. $\alpha$ captures the effective orientation of the active stress field generated by the helical muscles. Larger $\alpha$ increases the obliquity of the cable, thus increases the tendency of twisting as well as the tendency of the coupled system to generate multiple curvature reversals. As a result, the number of achievable sign-changing points along the trunk is increased.

On the other hand, $\beta$ governs the spatial arrangement of the effective anisotropy field. Varying $\beta$ changes how the two opposite-handed anisotropy fields superimpose and partially cancel, thereby altering bend–torsion decoupling and shifting the axial location of sign-changing points. Thus it shifts the axial placement of these sign-changing points and redistributes curvature along the trunk. In this view, shape complexity does not primarily emerge from concatenated segment-by-segment control, but from a low-dimensional geometric coupling law defined by $(\alpha,\beta)$. The illustration of the effect by varying $(\alpha,\beta)$ is shown in Figure~\ref{method2} and explained in Section~\ref{sec4} below.

\subsection{Biological interpretation as an effective anisotropy field}

Biologically, global activation denotes broad co-recruitment of left- and right-handed oblique muscle bands over a long axial extent. Local activation corresponds to sectoral or axially windowed recruitment that trims this global field. Under this interpretation, we hypothesize that the elephant may tune effective helical parameters through two complementary processes: selective global recruitment of left- and right-handed oblique muscle groups, which modulates the effective phase span $\beta$, and deformation-dependent changes in fibre orientation during trunk elongation or contraction, which modify the effective obliquity $\alpha$. 

In other words, the anatomical layout remains broadly fixed, while the effective helical mechanics vary through selective recruitment and deformation-dependent changes in fibre orientation. This leads to a dual-layer picture of control: much of the trunk’s posture backbone is generated by tuning an effective anisotropy field and corresponding functional helical parameters, while local recruitment as segmented activation can tune contacts, force application, and fine distal shaping.

\section{Minimal Evidence From A Simplified Prototype}\label{sec4}

To test whether dual-helicity coupling can in principle generate the required posture diversity, we built a cable-driven continuum robot prototype based on spiralling designs from \cite{huang2025grasping, wang2024spirobs}, equipped with opposite-handed helical tendons running from base to tip. Although simplified, this system serves as a proof-of-concept platform for testing how dual-helicity geometry can generate posture diversity under low-dimensional actuation.

\begin{figure*}[!t]
\centering
\includegraphics[width=\textwidth]{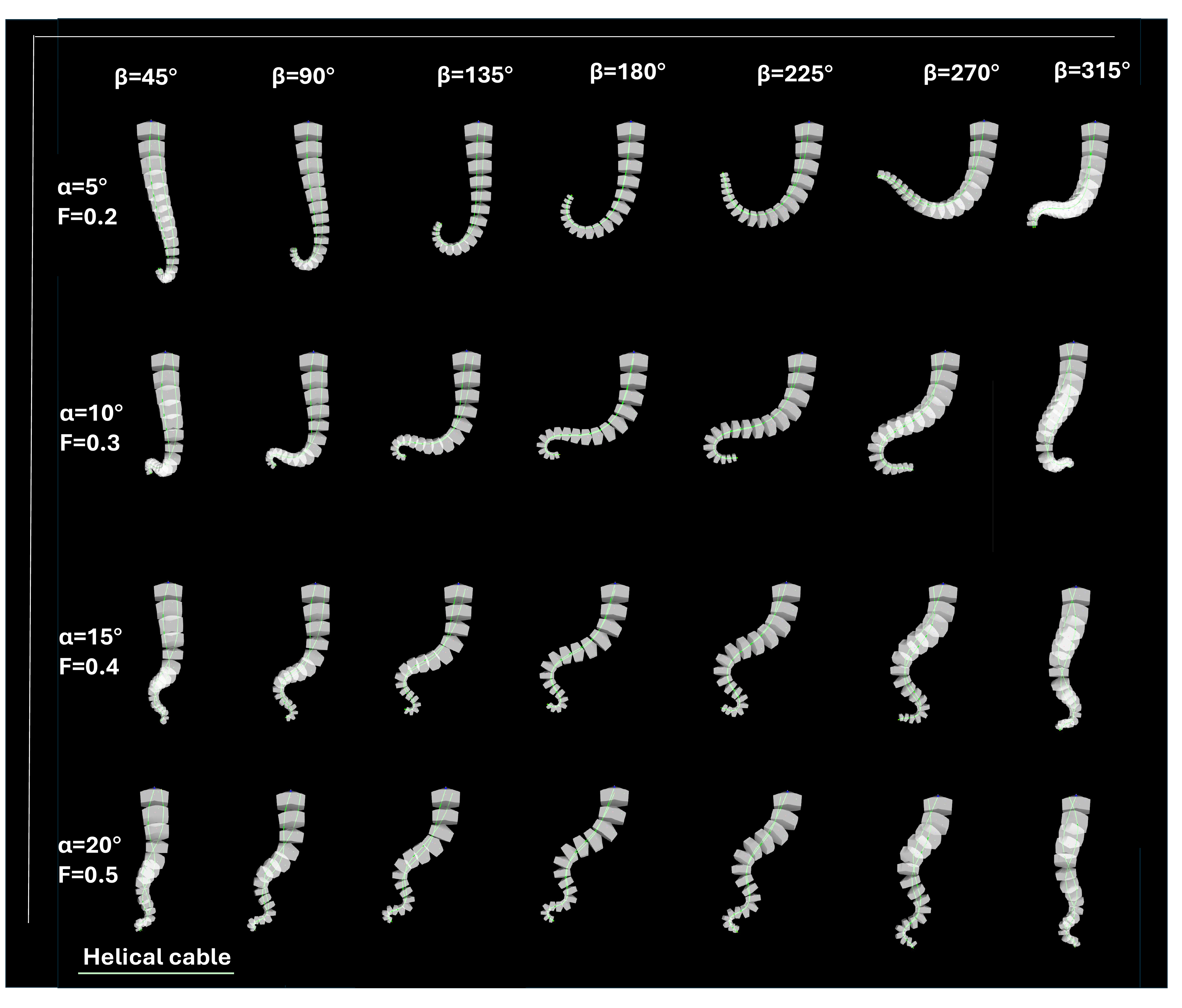} 
\caption{ A parameter sweep for $\alpha$ and $\beta$ is shown in MuJoCo simulation environment. The helix angle $\alpha$ was varied from $5^{\circ}$ to $20^{\circ}$ in $5^{\circ}$ increments, and the phase span $\beta$ from $45^{\circ}$ to $315^{\circ}$ in $45^{\circ}$ increments, under incrementing antagonistic actuation (F from $0.2 – 0.5$, normalized from $50 N$ per cable. Increasing F at fixed parameters only deepens existing bends and does not create new curvature sign-changes). The number of sign-changes increases non-decreasing with $\alpha$ for every tested $\beta$, confirming that obliquity governs shape complexity.  For a given $\alpha$, the sign-change position varies with $\beta$, consistent with the predicted role of phase span in redistributing curvature nodes.}
\label{method2}
\end{figure*}

Figure~\ref{method1} shows that by tuning only $\alpha$ and $\beta$, the prototype can reproduce a broad library of elephant-like postures relevant to grasping, including distal bends, hooks, twisting configurations, single-node S-bends, and multi-node reverse-bent shapes. Importantly, these postures emerge without concatenated segmental actuation. Rather than demonstrating a full biological reproduction of trunk mechanics, the prototype demonstrates that much of the required posture complexity can already arise from geometric coupling in a simplified system.  Figure~\ref{method2} and Table~\ref{table1} has demonstrated the trends of the parameter changes: The number of sign-changes increases non-decreasing with $\alpha$ for every tested $\beta$, confirming that obliquity governs shape complexity.  For a given $\alpha$, the sign-change position varies with $\beta$, consistent with the predicted role of phase span in redistributing curvature nodes.

\begin{table}[htbp]
\caption{Number of Curvature Sign-change For ($\alpha$, $\beta$) Combinations}
\label{tab:curvature_sign_change}
\centering
\footnotesize
\setlength{\tabcolsep}{5pt}
\begin{tabular}{c|ccccccc}
\hline
 & $45^\circ$ & $90^\circ$ & $135^\circ$ & $180^\circ$ & $225^\circ$ & $270^\circ$ & $315^\circ$ \\
\hline
$\alpha=5^\circ$  & 1 & 1 & 0 & 0 & 0 & 1 & 1 \\
$\alpha=10^\circ$ & 2 & 2 & 1 & 1 & 1 & 1 & 2 \\
$\alpha=15^\circ$ & 3 & 2 & 2 & 2 & 2 & 2 & 3 \\
$\alpha=20^\circ$ & 4 & 3 & 3 & 3 & 3 & 3 & 4 \\
\hline
\end{tabular}\label{table1}
\end{table}


This observation motivates a robotics insight as well as a biological one. In robotics, it suggests that high-DoF posture generation may not always require equally high-dimensional control, but can instead be partially offloaded to geometry-aware morphological design. In biology, it suggests that elephant trunks may exploit comparable geometric synergies, using low-dimensional co-recruitment to generate the backbone of posture while reserving more localized control for contact regulation and fine manipulation.

\textbf{Scope and Limits.} 
The prototype is not intended as a biomechanical replica of the elephant trunk, since it lacks muscular hydrostats, constant-volume constraints, and elongation/contraction capability.  It serves as a minimal physical testbed for isolating the geometric coupling effect and a cross-discipline thinking tool for formulating a hypothesis that needs further validation.

In reality, the oblique muscle groups are organized as discrete helical bands rather than continuous wraps. However, at a macroscopic scale their combined action can be approximated by a continuous helical sheath, an assumption commonly adopted in soft robot modeling \cite{hu2023bioinspired, leanza2024elephant, kaczmarski2024minimal}. Therefore, our helical cable design is a simplification of the oblique muscle groups, which preserves the essential torsion-bending geometric coupling while reducing anatomical complexity to a tractable form. The dual-helicity coupling serves as the underlying kinematic framework regardless of whether the actuation is driven by cables or hydrostat pressure.

Biologically, the helical muscles in real elephant trunks are organized in discrete groups, which allows the animal to selectively activate muscles at different phases (dynamically modulating $\beta$) or elongate/contract the trunk to alter the effective helical pitch (dynamically modulating $\alpha$). This dynamic modulation is not currently achievable in our simplified robotic prototype, where the helical muscle groups are abstracted into representative cables with fixed geometric parameters. Consequently, the kinematic versatility of any single hardware configuration is constrained. This specific hardware mismatch explains the necessity of our extensive parameter evaluation (e.g., Fig.\ref{method2}): since the robot cannot alter these parameters in real time, we must identify a single, optimally versatile fixed-parameter configuration that can adapt to a wide range of grasping scenarios. Nevertheless, even with fixed geometric parameters, this abstraction successfully demonstrates that embedding dual-helicity into morphology fundamentally reduces the required control dimensionality, offering an immediate benefit to the design of continuum robots.



Our account still emphasizes the importance of the segmental or local activation for the following facts: elephants can modulate grasping force independently of motion; the emergence of “pseudo-joint”; and the fine control of the trunk tip via the motor fovea. In the proposed dual-layer control strategy, low-dimensional global fields could set most of the posture on the first layer, while local trims tune contacts and force on the second layer.

Several physical effects not captured by our simplified prototype may become important at larger scales or under external loading. Gravity-induced sagging will bias the trunk's rest configuration and may shift curvature node locations. Tendon friction along the cable routing path may introduce history-dependent behavior \cite{wang2024exploiting}. External contact forces during grasping will locally perturb the posture away from the geometry-prescribed shape. We anticipate that these effects will modify the quantitative mapping from $(\alpha, \beta)$ to posture shape, but do not eliminate the underlying geometric coupling. Therefore, while the dual-helicity geometry dictates the macroscopic posture, deploying this framework in the real world will ultimately require pairing it with local closed-loop control or residual learning to bridge the gap between idealized geometry and physical hardware dynamics.

\section{Testable predictions and robotics insights}

Our dual-helicity hypothesis not only offers a biologically testable account of elephant trunk shape formation, but also suggests new design principles for continuum robotics. Rather than treating biology and robotics as separate domains, each prediction below links a biological question to a corresponding robotics insight.

\begin{enumerate}

\item \textbf{$\alpha$-controlled sign-change count.} Segments with greater obliquity should exhibit more curvature sign-changes under comparable tasks.

\textbf{Robotics insight:} Tuning helical obliquity provides a geometric route to generate multi-node reverse-bent shapes without increasing the number of independently controlled actuation channels.

\textbf{Biology test:} Estimate regional oblique-fibre angle using MRI or ultrasound elastography, and correlate it with the observed number of sign-changing points.

\item \textbf{$\beta$-steered node placement.} Different relative phase spans between opposite-handed oblique muscle groups should predict axial shifts in S-node placement.

\textbf{Robotics insight:} Relative phase offset provides a compact design variable for steering where curvature reversal occurs, suggesting a new way to program posture families without segmented architectures.

\textbf{Biology test:} Quantify the axial location of curvature reversal points in video or imaging data and compare them with inferred muscle distribution patterns.

\item \textbf{Energetic and kinematic economy.} For distal trunk alignment before grasping, strategies with S-bends should require less proximal bending and body displacement than strategies without sign-changing curvature.

\textbf{Robotics insight:} Sign-changing curvature may improve pre-grasp alignment while reducing proximal motion, suggesting a morphology-based strategy for energy-efficient reaching in continuum manipulators.

\textbf{Biology test:} Compare base angle, body displacement, and where feasible indirect metabolic proxies during controlled reaching tasks.

\end{enumerate}

These predictions suggest a concrete agenda for both robotics and biology. In robotics, dual-helicity offers a route toward geometry-aware continuum manipulators that generate rich posture vocabularies with reduced control dimensionality. In biology, advances in anatomical imaging and behavior quantification may help determine whether similar low-dimensional geometric shortcuts are used in the elephant trunk. In this sense, robotic simplifications do not replace biology; they provide experimentally accessible models for exposing its hidden geometric principles.

Finally, we note that dual-helicity geometry is complementary to data-driven and learning-based control approaches that are increasingly prevalent in soft robotics. In reinforcement learning, for instance, the choice of action space critically affects learning efficiency: a policy that must discover complex postures by independently controlling dozens of segments faces a high-dimensional search problem. Dual-helicity offers a physically grounded dimensionality reduction: by parameterizing the action space in terms of $(\alpha, \beta)$ rather than per-segment activations, the search space is compressed to a low-dimensional manifold that already contains the task-relevant posture families. In this sense, the geometric shortcut acts as a powerful structural prior that accelerates learned controllers. Local segmental adjustments can then be seamlessly integrated via learned residual policies or localized feedback to handle the fine manipulations that the global geometry does not capture.

\addtolength{\textheight}{-12cm}   




\section*{ACKNOWLEDGMENTS}
This research is supported by the Singapore MIT Alliance for Research and Technology (SMART) centre. This research is also supported by A*STAR, Singapore, through the Italy-Singapore collaborative project “DESTRO - Dextrous, strong yet soft robots”.

\bibliographystyle{IEEEtran}
\bibliography{root}

\end{document}